\documentclass[11pt]{article}
\usepackage{coling2020}
\usepackage{times}
\usepackage{url}
\usepackage{latexsym}
\usepackage{array}
\usepackage{ragged2e}
\usepackage{graphicx}
\usepackage{multirow}

\usepackage{color}

\colingfinalcopy 

\title{Incorporating Count-Based Features into Pre-Trained Models for Improved Stance Detection}

\author{Anushka Prakash \\
 School of Computer Science \\
  University of Birmingham\\
  United Kingdom \\
  {\small \tt anushkaprakash6@gmail.com } \\\And
Harish Tayyar Madabushi \\
  School of Computer Science \\
  University of Birmingham\\
  United Kingdom \\
  {\small \tt  harish@harishtayyarmadabushi.com} \\
}

\date{}

\begin{document}
\maketitle
\begin{abstract}
  The explosive growth and popularity of Social Media has revolutionised the way we communicate and collaborate. Unfortunately, this same ease of accessing and sharing information has led to an explosion of misinformation and propaganda. Given that stance detection can significantly aid in veracity prediction, this work focuses on boosting automated stance detection, a task on which pre-trained models have been extremely successful on, as on several other tasks. This work shows that the task of stance detection can benefit from feature based information, especially on certain under performing classes, however, integrating such features into pre-trained models using ensembling is challenging. We propose a novel architecture for integrating features with pre-trained models that address these challenges and test our method on the RumourEval 2019 dataset. This method achieves state-of-the-art results with an F1-score of 63.94 on the test set.
  \end{abstract}

\section{Introduction and Motivation}
\label{intro}
%
%
\blfootnote{
    %
    %
    %
    %
    \hspace{-0.65cm}  
    Accepted for publication to the Third Workshop on NLP for Internet Freedom (NLP4IF2020) \\
    \hspace{-0.65cm}  
    This work is licensed under a Creative Commons 
    Attribution 4.0 International Licence.
    Licence details:
    \url{http://creativecommons.org/licenses/by/4.0/}.
    %
    %
}

In today's world, it is hard to imagine life without the internet and social media. The exponential growth of social media platforms over the past decade has provided people with an extremely convenient medium to access, consume and disseminate information. Social media has also become the primary source of news due to its ability to allow users to follow real-time updates \cite{10.1145/2675133.2675202}. Unfortunately, this has also led to the spread of misinformation and propaganda. A recent example is that of Covid-19 which saw a plethora of misinformation online \cite{Donovon_2020}.

Social media users often have trouble identifying credible resources on various platforms and are frequently guided towards misinformation, causing users to trust misinformation. With about 200 billion tweets posted every day \cite{krawczyk2017sentiment}, manually processing this information to detect rumour can be extremely tedious, time consuming, prone to human error and bias and most importantly impractical. Thus, it is essential to develop efficient automated methods that help in detecting and tracking misinformation online to aid in the fight against misinformation.

There are four primary components to rumour determination: rumour detection, rumour tracking, stance classification, and veracity classification \cite{Zubiaga2018DetectionAR}. Stance classification aims at studying the stance of a reply post (referred to as an `Opinion') on the source post (referred to as the `Target' - to what the opinion is subjected). Studies have shown that posts containing false information have a notably higher number of `denying’ replies than the ones that do not \cite{derczynski2015pheme}. Hence, stance classification can play an important role in assessing the veracity of a rumour. 

Given that stance classification can strongly aid in determining the veracity of a post, it is an important if challenging problem to solve. It remains challenging because posts are often sarcastic or witty, while also questioning the validity of the original poster's assumptions or evidence~\cite{hasan-ng-2013-stance}.

\subsection{Count-based Features and Pre-Trained models}
\label{handcraft-feats}

Recent trends in Natural Language Processing (NLP) have seen the rise of pre-trained language models like GPT \cite{radford2018improving}, BERT \cite{devlin-etal-2019-bert}, RoBERTa \cite{Liu2019RoBERTaAR} and so on. These models are pre-trained on corpora containing millions of words and can be fine-tuned on several tasks to achieve state-of-the-art results. However, it is conceivable that these deep models can benefit from handcrafted features based on human intuition such as an extensive error analysis of these systems' results. For example, early work by \newcite{10.5555/1860631.1860645} on ideological stance detection observed that a simple unigram model outperformed more complex models using sentiment. This highlights that the stance of a statement is often related to the specific words used to express that opinion (for example, words such as `Fake’, `Not true’, `disagree’ are often used to deny a target claim), and stance classification systems could benefit from explicit features depicting such words. This is especially true when models are used to analyse social media data where elements such as hash tags could have special meaning that can be incorporated using handcrafted features. Furthermore, feature engineering can often be used to boost performances on specific classes that deeper models fail to perform well on. Hence, while it is evident that pre-trained language models have outperformed classical neural networks, simple models trained on hand crafted features can perform well on specific classes or instances that deeper models fail on.

Not only can models designed to analyse social media benefit from the use of handcrafted and count-based features but they are particularly important in handling disinformation and fake news as knowledge of the world at large and global events is required for improved tracking of disinformation. This non-linguistic information is likely to be available through count-based and handcrafted features but not in the  pre-trained vector. This is also supported by prior work by \newcite{Aker2017SimpleOS} and \newcite{bahuleyan-vechtomova-2017-uwaterloo}, discussed in Section \ref{related-work}.

Given that handcrafted and count-based features can be used to improve the performance of deep learning models on specific classes or kinds of input, it is natural to ensemble these simple models based on handcrafted features with deeper models such as BERT. However, this is rather challenging for two important reasons: 
\begin{enumerate}
     \setlength\itemsep{0.01em}
    \item Pre-trained models, such as BERT, are often trained for between 2 and 5 epochs during fine-tuning whereas simpler feature based models need to be trained for much longer. Our experiments show that a simple ensemble of these models results in over-fitting (Section \ref{explore-exp}) . 
    \item There are likely to be too many features to directly ensemble the raw features with pre-trained models (resulting in too much noise), a loss of important - task specific - information when using dimensionality reduction methods (Section \ref{explore-exp}), and too few output classes to use only the outputs of a feature based model in an ensemble (lack of information).
\end{enumerate}

Therefore, this paper focuses on establishing methods of effectively integrating count-based features into pre-trained models such as BERT and RoBERTa, with specific regard to the identification of disinformation and propaganda on social media, where handcrafted and count-based features can be particularly helpful.

In this work, we formulate stance classification as a multi-class classification problem and experiment with various state-of-the-art NLP techniques to establish the best method of achieving this objective. We use the dataset provided for RumourEval 2019 (Task 7a of SemEval 2019) or the task of stance classification \cite{gorrell-etal-2019-semeval}. While tweets have been extensively studied in the past due to their vast reach and brief nature \cite{krawczyk2017sentiment}, this data set also includes posts from Reddit, so focusing on a broader range of social media data. Each conversation thread consists of a source post which states or discusses a rumour. This source post is followed by a set of replies (and replies to these replies) which either `Support', `Deny', `Query' or `Comment' on the source text (See Figure \ref{fig:dataset-example-fig}). The source post is also labelled depending on the stance it takes at its (hidden) target~\cite{gorrell-etal-2019-semeval}.
 
So as to ensure reproducibility and enable other researchers to build upon this work, we make our program code, hyperparameter details and models available~\footnote{\url{https://github.com/Anushka-Prakash/RumourEval-2019-Stance-Detection/}}.

\section{Related Work}
\label{related-work}

The effectiveness of stance detection in identifying and tracking misinformation online has led to significant research in this area. This section provides an overview of work related to this study.  

Early work on studying perspectives such as that by \newcite{lin-etal-2006-side} use Naïve Bayes and SVM based classifiers to show that word usage alone can help in determining the point of view of a document or sentence. In a similar study, \newcite{kim-hovy-2007-crystal} go beyond word usage and exploit the lexical patterns used by people while expressing an opinion.~\newcite{10.5555/1860631.1860645} showed that a unigram based model outperformed a sentiment-based model for determining stance thus reinforcing the importance of words in detecting stance. This shows that the stance of an opinion could be related to the kind of words used to express it. Although working on a different problem set (focused on binary classification) and using different datasets, the above-mentioned studies are relevant to this work because they emphasise on the importance of words used to express an opinion to determine their stance.

\newcite{bahuleyan-vechtomova-2017-uwaterloo} hypothesise the presence of certain words in the reply text or the opinion as an indicative measure of its stance. They use these and some tweet specific features like the word count of the tweet to train on a gradient boosting classifier and show that stance classifiers could benefit from these topic independent features. \newcite{Aker2017SimpleOS} explore a novel set of features and claim that simpler classifiers with profuse feature knowledge can outperform several complex machine-learning techniques in stance detection. \newcite{riedel2017simple} use a Multi Layer Perceptron (MLP) having a single hidden layer with term frequency and TF-IDF vectors as inputs to perform at par with several complex models. These works lend weight to our hypothesis that handcrafted and count-based features can significantly aid in the task of stance detection. 

Recent work in stance classification shows extensive use of deep-learning models to achieve state-of-the-art results.
\newcite{fajcik-etal-2019-fit} and \newcite{yang-etal-2019-blcu} who worked on the same task and dataset as this work, use the BERT and GPT architectures respectively for stance classification. \newcite{fajcik-etal-2019-fit} use an ensemble model wherein they combine outputs from several BERT models to increase the F1-score. Their system achieves a significantly high F1-score without the use of any hand-crafted features. \newcite{yang-etal-2019-blcu}, who achieved state of the art results on this dataset prior to this work, use an inference chain based system that was fine-tuned on GPT. They use handcrafted features such as the presence of question marks, URLs, positive and negative words, etc. and leverage the structure of a conversation thread to achieve state-of-the-art results in stance classification.

\subsection{Class-imbalance}
\label{class-imabalance-lr}

Since most data sets based on real-world data (including the stance detection data set used in these experiments) suffer from class-imbalance, we explore recent methods of addressing this issue. One method of addressing class imbalance is the use of a two-step classifier, such as work by \newcite{wang-etal-2017-ecnu}. They first classify the tweets as `comment’ and `non-comment’ and use a second classifier to distinguish non-comments as `Support’, `Deny’, or `Query’. \newcite{krawczyk2017sentiment}, on the other hand, propose a one-vs-one decomposition of the multi-class problem before then combining the outputs of these binary classifiers using a weighted approach to rebuild the multi-class problem. \newcite{yang-etal-2019-blcu}, on the other hand, use examples from similar data sets to increase the training data for minority classes.

The other method of handling class-imbalance is by use of cost-weighting. \newcite{tayyar-madabushi-etal-2019-cost} apply cost-sensitive learning for the task of Fine-Grained Analysis of Propaganda in News Article ~\cite{da-san-martino-etal-2019-fine} wherein they first test the similarity between training and validation sets using Wilcoxon signed-rank test. They also show that cost-weighting (increasing the cost of a minority class) can be more beneficial when applied to dissimilar datasets. \newcite{fajcik-etal-2019-fit} also use cost-weighting to tackle class-imbalance. 

\section{Experimental and Model Design}
\label{exp-design}

This section describes initial exploratory experiments and the process behind building a model that address the two obstacles to integrating features into pre-trained models mentioned in Section \ref{handcraft-feats}, namely: \textbf{a)} the significant difference in the number of epochs required by pre-trained models and feature based models, and \textbf{b)} the fact that using raw features might contain too much noise in an ensemble, the outputs alone might contain too little information, and dimensionality reduction methods result in the loss of task specific information. 

\subsection{Pre-Processing and Experimental Setup}
\label{pre-processing}

The pre-processing steps used for all our experiments were consistent. URLs and mentions present in all the source and the reply posts were replaced with the special tokens `\$URL\$' and `\$MENTION\$' respectively. This was adapted from \newcite{fajcik-etal-2019-fit} and found to be helpful in our experiments. 

We use two sequences as input to pre-trained model, which represent an opinion (the 1\textsuperscript{st} sequence) and its target (the 2\textsuperscript{nd} sequence). It is the 1\textsuperscript{st} sequence, the opinion, that we are interested in classifying. This first sequence consists of the reply being classified and, in instances where such a post is itself a reply to another reply, is concatenated with its `parent'. Figure \ref{fig:dataset-example-fig} illustrates the tree structure of a conversation thread from the dataset. 

Inputs to the pre-trained models were generated as follows: \textbf{(1)} While classifying the source post of a conversation thread (TE-1 in Figure \ref{fig:dataset-example-fig}), the text of this source post is the first input sequence (the one being classified). Since this post is not a reply and the target information is not explicitly available, the second input sequence is empty. \textbf{(2)} When classifying a reply (for example, TE-2 in Figure \ref{fig:dataset-example-fig}), the text of this reply is the first input sequence and the source post of the conversation thread becomes the second sequence. \textbf{(3)} While classifying a nested reply (reply to a reply, such as TE-3 in Figure \ref{fig:dataset-example-fig}), the text of this nested reply is concatenated with the text of its parent and used as the first input sequence, and the source post of the entire conversation thread becomes the second sequence. This method of structuring the input was adapted from work by \newcite{fajcik-etal-2019-fit} who hypothesise that the stance of a given post depends on itself, its previous post and the source post of the conversation thread. 

We use the \emph{encode\_plus} function provided by HuggingFace in their re-implementation of RoBERTa to encode the input sequences. The input sequences are first appended with special tokens - RoBERTa uses special tokens to indicate the start of the input ($<s>$), the end~($</s>$) and two end tags to differentiate between different input sequences ~($</s></s>$). Figure \ref{fig:dataset-example-fig} illustrates how special tokens were embedded in each example. These sequences are then encoded using Byte-Pair encoding and fed to RoBERTa. 

\subsection{Exploratory Experiments}
\label{explore-exp}

Initial experiments using BERT yielded an F1-score of 0 for the class `Deny’, one of the harder to predict labels (See table \ref{exploratory-exp-results-table}). RoBERTa, however, produced an F1-score greater than 0 for all 4 classes, which led us to pick RoBERTa over BERT. We pre-process each example and feed two sequences as input to RoBERTa (as explained in section \ref{pre-processing}).

\begin{figure}[ht]
    \centering
    \begin{tabular}{c}

    \noindent\fbox{%
    \parbox{0.9\textwidth}{%
        \textbf{Source Post (TE-1)}: Darren Wilson is a six year veteran of the \#Ferguson Police and had no disciplinary actions against him. \textbf{[Support]}  \vspace{1mm}\par
         \hspace{5mm}
         \textbf{Reply 1 (TE-2)}: Can we see video proof \textbf{[Query]} \vspace{1mm}\par
        \hspace{5mm} \textbf{Reply 2 (TE-3)}: HE ISN'T THE SHOOTER RT [MENTION] \textbf{[Comment]} \vspace{1mm}\par 
        \hspace{10mm} \textbf{Reply 2.1 (TE-4)}: [MENTION] well who is \#Ferguson \textbf{[Comment]}
        \vspace{2mm}
        \hrule
        \vspace{5mm}
        
         \textbf{TE-1}: $<s>$ Darren Wilson is a six year veteran of the \#Ferguson Police and had no disciplinary actions against him. $</s>$ $</s>$ $</s>$  \vspace{5mm}\par
        
        \textbf{TE-2}: $<s>$ Can we see video proof $</s>$ $</s>$ Darren Wilson is a six year veteran of the \#Ferguson Police and had no disciplinary actions against him. $</s>$ \vspace{5mm}\par

        \textbf{TE-3}: $<s>$ HE ISN'T THE SHOOTER RT [MENTION] $</s>$ $</s>$ Darren Wilson is a six year veteran of the \#Ferguson Police and had no disciplinary actions against him. $</s>$ \vspace{1mm}\par 
        
        \textbf{TE-4}: $<s>$ [MENTION] well who is \#Ferguson HE ISN'T THE SHOOTER RT [MENTION] $</s>$ $</s>$ Darren Wilson is a six year veteran of the \#Ferguson Police and had no disciplinary actions against him. $</s>$ 
        
    }%
    }
        \end{tabular}
    \caption{An example of a conversation thread and associated labels (in bold at the end of each post) from the RumourEval data set showing the tree like structure. The second part shows the training examples (TE) constructed for use in pre-trained models from each of the original posts. Further details in text.}
    \label{fig:dataset-example-fig}
\end{figure}

Next, we attempt to find a set of features that help in stance classification. As discussed in Section~\ref{handcraft-feats}, work by \newcite{10.5555/1860631.1860645} on ideological stance classification showed that a unigram model can outperform more complex models. While their study was on a different problem set and focused on binary classification (`for’ and `against’) of ideological stance, they showed that the stance of an opinion could be related to the kind of words used to express it. This is particularly true of a the `Deny’ class in the data set we work with.  As discussed in section \ref{handcraft-feats}, words such as `Not true’ and `Disagree’ are frequently used to deny a claim.

We train a Multi Layer Perceptron (MLP) using TF-IDF features as input. We use a single hidden layer consisting of 128 hidden units, a tanh activation function and a final linear layer with a softmax function to make predictions. We use a learning rate of 0.02 and train this model for 55 epochs (Additional  details such as the complete list of hyperparameter  are available as part of the program code released). This model achieves an F1-score on `Deny’ class greater than that achieved using the RoBERTa-base (See table \ref{exploratory-exp-results-table}).

These results indicate that the use of TF-IDF features was particularly beneficial in classifying `Deny’. In their recent work, \newcite{Lim2020UoBAS} incorporate TF-IDF features with BERT to improved performance on Offensive Language Identification in Social Media~\cite{zampieri2020semeval2020}. This shows that tasks like stance classification and abuse detection greatly benefit from information pertaining to the words used in expressing that emotion.

\begin{table}[!ht]
\begin{center}
\begin{tabular}{|> \justify m{4.2cm}|> \justify m{1.5cm}|c|c|c|c|}

\hline \textbf{Model} &  \textbf{Macro F1-Test}  & \textbf{F1 Support}  & \textbf{F1 Deny}  & \textbf{F1 Query}  & \textbf{F1 Comment}  \\ \hline

 MLP Model & 0.38 & 0.18 & 0.24 & 0.22 & 0.87 \\ 

BERT  & 0.46 & 0.44 & 0.0 & 0.52 & 0.90 \\ 

RoBERTa-base  & 0.51 & 0.46 & 0.14 & 0.52 & 0.91 \\ 

RoBERTa-base + All TF-IDF Features & 0.49 & 0.45 & 0.06 & 0.53 & 0.91 \\ 

RoBERTa-base + PCA Transformed TF-IDF 
 & 0.56 & 0.34 & 0.46 & 0.54 & 0.89 \\

RoBERTa-base + Output of MLP  
 & 0.51 & 0.46 & 0.19 & 0.50 & 0.91 \\\hline
 
\end{tabular}
\end{center}
\caption{\label{exploratory-exp-results-table} Achieved results on the test set for Exploratory Models}
\end{table}

Given the improved performance of the MLP on `Deny', we aim to create an ensemble of the MLP and a pre-trained model so as to boost the overall performance. We ensemble the pooled output of RoBERTa with the output of a Multi-Layered Perceptron (MLP) (consisting of 4 units, one for each class). This combination is then connected to a linear layer followed by a softmax function to make predictions. While this model performed slightly better on the `Deny' class, it performed worse on the class `Query', thus achieving an overall Macro F1 score exactly equal to RoBERTa Base. We also create an ensemble model combining all the TF-IDF features with RoBERTa which we train for 4 epochs. This model performs worse than RoBERTa-base. 

In order to verify that this decrease in performance is related to the TF-IDF features being high dimensional and noisy, we conduct experiments wherein we apply Principal Component Analysis (PCA), to reduce the dimensions of the TF-IDF vector to a length of 128 before using this shortened vector in the ensemble. We train this ensemble model for 40 epochs and note an increase in performance over the first 4 epochs and a subsequent increase in loss (with no change in accuracy) indicating that the model overfits. We similarly train the ensemble of RoBERTa and the MLP for 20 epochs and find that the model dramatically overfits with a drop in F1 from 0.51 to 0.15. 

This section described experiments on six different models (See Table \ref{exploratory-exp-results-table} for comparative results, we use the macro-averaged F1-score to compare models). Our finds are that \textbf{a)} the use of an MLP can be beneficial in boosting performance, \textbf{b)} the use of all tf-idf features in an ensemble is too noisy, \textbf{c)} the output of the MLP alone contains too little information to be useful in an ensemble, and \textbf{d)} ensembles of pre-trained models and an MLP will suffer from over-fitting before the MLP can be trained sufficiently. Additionally, training an ensemble of an MLP and RoBERTa takes about 10 min per epoch thus dramatically increasing the training time when we train for 20 or 40 epochs.  

We address these shortcomings by introduction of a novel architecture that makes use of an \textit{already trained MLP} which prevents overfitting, reduces train time  while also ensuring the effective integration of pre-trained models and handcrafted features. The next Section (\ref{ensemble-model}) describes this model.

\subsection{Model Architecture}
\label{ensemble-model}

This section describes the novel ensemble architecture that was used to incorporate count-based features, particularly useful for the classification of posts belonging to `Deny', while simultaneously addressing the difficulties in ensembling feature based models and pre-trained models. 

\begin{figure}[h]
\begin{center}
  \includegraphics[width=0.9\linewidth]{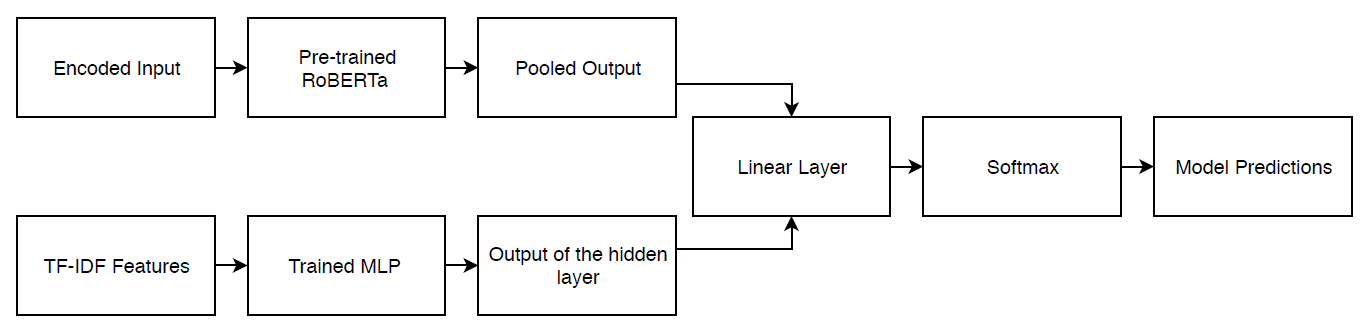}
  \caption{\label{fig:proposed-architecture}Architecture of the Proposed Model which combines output from RoBERTa and the hidden layer of an already trained MLP model. }
  \end{center}
\end{figure}

Figure \ref{fig:proposed-architecture} provides an illustration of the proposed architecture. This ensemble architecture uses two novel elements to address the two difficulties in creating an ensemble of the feature based and pre-trained models. Specifically \textbf{a)} the MLP used in the ensemble is one that \textit{is already trained and optimised} (hyperparameters) for this task, and \textbf{b)} the output of the \textit{hidden layer} of this already trained MLP is used in ensembling with pre-trained models instead of either the input or the final output. 

The use of an already trained MLP in the ensemble ensures that the MLP does not underfit when trained in an ensemble with pre-trained models for a small number of epochs. This method, while similar to pre-training is not exactly the same as it is trained on the same task unlike in pre-training. The use of the hidden layer in the ensemble simultaneously deals with the problem of having too much noise in the input and too little information in the output by providing an abstract layer of condensed information from the MLP as input to the final ensemble model.

More concretely, this ensemble model is constructed as follows: The pooled output from RoBERTa (vector of length 768) is concatenated with the output of the hidden layer (vector of length 128) of an already trained MLP. This combined vector, having a length of 896, is further connect to a linear layer and a softmax function to make predictions (See Figure \ref{fig:proposed-architecture}). We train this ensemble model for 6 epochs with a learning rate of 2e$^{-6}$ and batch size 4. Further details on this model are available in the program code released as part of this work. We also test on ensembling the output of an already trained MLP instead of the hidden layer, and while it does do better than RoBERTa alone, it does not do as well as an ensemble of the hidden layer.

This ensemble model outperforms RoBERTa along with all previously used ensemble models. We find this result, presented in Table \ref{ensemble-results-table}, to be consistent across both RoBERTa-base and RoBERTa-large.

To address the possibility that our model was caught in local optima, we train each of our models 5 times using different seeds and evaluate them using the development set. We then use the model that performs the best on the development set to classify the test set and report results.

\subsection{Tackling Class-imbalance}
\label{tackling-class-imb}

Class-imbalance implies that the number of training examples of one or more classes are significantly lower than the other classes. The RumourEval data set suffers from this same problem (Support 13.9\%, Deny 6.6\%, Query 4.8\% and Comment 72.4\%), and the `Comment’ class is not only dominant in the data set but is also the least helpful in detecting rumour \cite{gorrell-etal-2019-semeval}. While there are several ways of handling class-imbalance (as discussed in section \ref{class-imabalance-lr}), we make use of cost-weighting due to its flexibility and proven performance.

Cost-weighting consists of assigning a higher weight to a minority class which holds more relevance. The cost function is thus modified to accommodate the weights assigned to each class - the more the weight, the more the model is penalised for inaccurate predictions on that class. The cost weights for our models were empirically determined via experiments and significantly boosted performance. We incorporate cost-weighting in all our experimented models.

\section{Results and Analysis}
\label{results-and-analysis}

This section presents an overview of our experiments and findings. We compare our results with prior state-of-the-art work on stance classification which makes use of this data set. We also perform an error analysis of our proposed model and report findings. 

\subsection{Results}
\label{results}

Table \ref{ensemble-results-table} presents the macro-averaged F1-score we obtained by our proposed model (an ensemble model combining RoBERTa's pooled output with the output of the hidden layer of an already trained MLP). We also present results from both RoBERTa-base and RoBERTa-large to draw attention to how our proposed method significantly improves performance on both these versions of RoBERTa. For each experiment, we also report the macro-averaged F1-score and the F1-scores for all classes. These results show how how our proposed architecture significantly improves accuracy on the previously under-performing class `Deny', which our count-based features were designed to improve.

\begin{table}[h]
\begin{center}
\begin{tabular}{|> \justify m{4.8cm}|> \justify m{1.5cm}|c|c|c|c|}

\hline \textbf{Model}  &  \textbf{Macro F1-Test} & \textbf{F1 Support} & \textbf{F1 Deny}  & \textbf{F1 Query} & \textbf{F1 Comment}  \\ \hline

RoBERTa-base & 0.51 & 0.46 & 0.14 & 0.52 & 0.91 \\  

RoBERTa-base + Hidden layer output of Trained MLP
 & \textbf{0.58} & \textbf{0.43} & \textbf{0.39} & \textbf{0.58} & \textbf{0.92} \\  \hline
 
RoBERTa-large
 & 0.57 & 0.43 & 0.42 & 0.54 & 0.92 \\  
 
RoBERTa-large + Hidden layer output of Trained MLP
\textbf{(Proposed Model)} 
 & \textbf{0.64 } & \textbf{0.48} & \textbf{0.55} & \textbf{0.60} & \textbf{0.93} \\  \hline

\end{tabular}
\end{center}
\caption{\label{ensemble-results-table} Results of our Proposed architecture on the test set using  RoBERTa-base and RoBERTa-large showing a consistent improvement in performance across both versions of RoBERTa.}
\end{table}

Table \ref{comparative-study-table} presents our results alongside top performing systems on this dataset. We compare our results with the winning teams of SemEval 2019 for the same task. A branch-LSTM based system was used as a baseline for the stance classification task of SemEval 2019. We note that the model proposed in this work achieves state-of-the-art results on this dataset.

\begin{table}[!h]
\begin{center}
\begin{tabular}{|> \justify m{3.8cm}|m{5cm}|c|}
\hline \textbf{System} & \textbf{Details} & \textbf{Macro Averaged F1-Score (\%)} \\ \hline
\textbf{This work} & & \textbf{63.94} \\[2ex]
\newcite{yang-etal-2019-blcu}  & 1\textsuperscript{st} Rank - SemEval 2019 & 61.87 \\
\newcite{fajcik-etal-2019-fit} & 2\textsuperscript{nd} Rank - SemEval 2019 & 61.67  \\ [1ex]
\newcite{kochkina-etal-2017-turing} & Baseline - SemEval 2019 (Winning method, SemEval 2017) & 49.3  \\
\hline
\end{tabular}
\end{center}
\caption{\label{comparative-study-table} A comparison of this work with prior state-of-the-art methods on the same task and data set.}
\end{table}

\subsection{Discussion}
\label{discussion}

Previous sections provided details of the various models we experimented with, along with their results on the test set. In this section we present an overview of our intuitions behind and observations from these experiments.

Our initial experiments showed how a feature based model can outperform pre-trained models \emph{on specific classes}. This is a rather intuitive result which could potentially be true of various tasks that benefit from handcrafted and count-based features. We reiterate that tasks related to disinformation, propaganda and rumour are particularly capable of benefiting from handcrafted and count-based features such as user information, hashtags, URLs, word frequencies, and so on, as these features provide additional information that can be useful in classification and possibly unavailable to pre-trained models pre-trained on corpora that are different from social media and fine-tuned on the task over very few epochs.

We then presented two significant challenges (Section \ref{explore-exp}) that are faced in creating an ensemble of features and pre-trained models such as BERT or RoBERTa. We addressed the first challenge of too many or too few input features to be combined with pre-trained models by using the output of the MLP's hidden layer. We then address the second challenge - the difference in the number of training epochs required by feature-based and pre-trained models - by use of an already trained MLP. 

The experiments with the ensemble model using RoBERTa’s pooled output and the output of the trained MLP model showed a significant increase in model performance on stance classification. We show how we address each challenge through our experiments and propose a novel architecture that utilises abstract information from a trained MLP to increase the performance of pre-trained models.

In summary, this work shows that stance classification systems benefit from features that depict the kind of words used to express an opinion by ensembling TF-IDF features with RoBERTa. We also propose a novel architecture to effectively integrate a feature engineered model with pre-trained deep learning models so as to significantly boost performance on tasks that benefit from handcrafted features (such as stance classification).

\subsection{Error Analysis}
\label{error-analysis}

We perform an extensive error analysis of our proposed model. First, we study the impact of availability of trained features to RoBERTa for stance classification. As depicted from the confusion matrix (Table \ref{confusion-matrix-table}), the ensemble model using the abstract features from the trained MLP model specifically improves the performance of `Support', `Deny' and `Query' classes. Our model performs exceedingly well on the `Deny’ class, which previous models have typically struggled with. However, this comes at the cost of misclassifying some other classes (`Support’ and `Comment’) as `Deny’, it increases the overall F1-score. This observation is detailed further using examples from the data set.

\begin{table}[h]
\begin{center}
\begin{tabular}{|c|c|c|c|c|}
\hline
\multirow{2}{*}{True label} & \multicolumn{4}{c|}{Predicted label} \\

\cline{2-5}
 & Support & Deny & Query & Comment \\
\hline
 Support & \textbf{53 (46)} & 4 (1) & 1 (1) & 99 (109) \\
\hline
 Deny & 1 (1) & \textbf{52 (31)} & 3 (2) & 45 (67) \\
\hline
 Query & 14 (11) & 6 (4) & \textbf{52 (47)}  & 21 (31) \\
\hline
 Comment & 1 (1) & 26 (12) & 22 (31) & \textbf{1427 (1432)} \\
\hline

\end{tabular}
\end{center}
\caption{\label{confusion-matrix-table} Confusion matrix for RoBERTa-large and RoBERTa-large + Trained MLP models. Values associated with RoBERTa-large are in brackets. }
\end{table}

An analysis of the predictions made by the model on the test set shows that it learns to associate words such as `Fake’, `not’ and `not true’ with the `Deny’ class.  Again, this shows that effective integration of features by use of trained MLP models can significantly help in the classification of output classes that benefit from such features. While this was helpful in the classification of most examples from the `Deny’ class, some posts that used the aforementioned words were `comments’ and misclassified as `Deny’ (Refer Note 1 from Table \ref{error-analysis-table}). Similarly, some `comments’ that contained interrogative words and question marks were classified as `Queries’ without taking the content of the post into consideration (Refer Note 2 from table \ref{error-analysis-table}). This is interesting as it reintroduces the problems encountered by earlier models - problems that were, to some extent, addressed by contextual pre-trained models, showing the need for a careful balance between these two approaches. 

\begin{table}[h]
\begin{center}
\begin{tabular}{|c|> \justify m{4cm}|> \justify m{4cm}|c|c|}
\hline \bf Note & \bf Reply-post & \bf Target & \bf Predicted label & \bf True label\\ \hline
1 & \textit{\$MENTION\$ If this is fake, the poster should be charged with spreading mass panic} & \textit{This is crazy! \#CapeTown \#capestorm \#weatherforecast \$URL\$} & Deny & Comment\\
2 & \textit{Why would I need a kettle?} & \textit{Is it true most Americans don't own a kettle? If so, why not?} & Query & Comment\\
\hline
\end{tabular}
\end{center}
\caption{\label{error-analysis-table} Error Analysis of predictions by the RoBERTa + trained MLP ensemble model on the test data. While the model learns to associate words like `Fake' and `why' with the `Deny' and `Query' classes respectively, it gives too much weight to these features, resulting in errors. }
\end{table}

The model also finds it challenging to distinguish between a `support’ and a `comment’ with respect to a post - a problem that could be attributed these two classes being somewhat similar. The model is also not able to classify posts that contain a URL, possible due to the missing information from the URL. Finally, the model also often inaccurately predicts the labels of the source post. We believe that this is due to the explicit  unavailability of target information.

\section{Conclusions and Future Work}
\label{conclusions}

This work introduced a novel architecture that makes use of abstract information from an already trained MLP to boost RoBERTa on the task of stance detection. Our technique achieves state-of-the-art results with an accuracy of 86.69\% and a macro-averaged F1-score of 63.94 on a standard data set. We conclude that an effective integration of features with models such as RoBERTa can significantly increase model performance on tasks that benefit from such features and that our architecture is an effective solution.

In future work we aim to integrate entire conversational threads to study its impact of such information as, in this work, we limit the information pertaining to one example to reply, previous and the source post of the conversation thread.  We also aim to use these methods on other related propaganda and abuse datasets along with exploring the impact of other handcrafted features. 

\section*{Acknowledgements}

We would like to thank the NVIDIA Deep Learning Institute for the provision of AWS credits which we used to access GPU resources in this work. 

\bibliographystyle{coling}
\bibliography{coling2020}

\end{document}